\documentclass{article}
\usepackage{spconf,amsmath,graphicx}

\title{PREDICTING TONGUE MOTION IN UNLABELED ULTRASOUND VIDEOS USING CONVOLUTIONAL LSTM NEURAL NETWORKS}

\name{Chaojie Zhao$^{1}$,
	Peng Zhang$^{2}$,
	Jian Zhu$^{3}$,
	Chengrui Wu$^{4}$,	
	Huaimin Wang$^{5}$,
	Kele Xu$^{5}$\sthanks{Corresponding author.}
}

\address{$^1$Graduate School at Shenzhen, Tsinghua University, Shenzhen, China\\
	$^2$University of Science and Technology of China, Suzhou, China\\
	$^3$Department of Linguistics, University of Michigan, Ann Arbor, USA\\
	$^4$Columbia University in the City of New York, New York, USA\\
	$^5$National University of Defense Technology, Changsha, China\\
	 kelele.xu@gmail.com\\
}

\begin{document}
%
\maketitle
\begin{abstract}
A challenge in speech production research is to predict future tongue movements based on a short period of past tongue movements. This study tackles speaker-dependent tongue motion prediction problem in unlabeled ultrasound videos with convolutional long short-term memory (ConvLSTM) networks. The model has been tested on two different ultrasound corpora. ConvLSTM outperforms 3-dimensional convolutional neural network (3DCNN) in predicting the 9\textsuperscript{th} frames based on 8 preceding frames, and also demonstrates good capacity to predict only the tongue contours in future frames. Further tests reveal that ConvLSTM can also learn to predict tongue movements in more distant frames beyond the immediately following frames. Our codes are available at: https://github.com/shuiliwanwu/ConvLstm-ultrasound-videos.
\end{abstract}
\section{Introduction}
\label{sec:intro}
Speech articulation is characterized by coupled, transient and highly variable movements of various articulators. Despite the lack of invariance, tongue movements during speech can be predictable due to speaker-specific strategies and constraints imposed by the physical properties of the tongue and phonological structure of the language \cite{green2003tongue,kent1972cinefluorographic,takemoto2001morphological}. Understanding the dynamics of tongue motions not only is the central theoretical quest in speech production research but also inspires many potential practical applications, such as articulatory synthesis, silent speech interfaces and pronunciation training interfaces. 

Tongue movements during speech production can be captured by the ultrasound imaging technique. Ultrasound tongue imaging has been widely used in speech production and clinical research, as it provides a fast and non-invasive means to visualize the real-time movements of the tongue \cite{stone2005guide}. Extracting meaningful articulatory information from ultrasound image sequences remains a challenging task, though various methods have been proposed for contour tracking (e.g., \cite{li2005automatic, xu2016robust,xu2017speckle}) and whole image analysis (e.g., \cite{JI201842,hueber2007eigentongue,xu2017convolutional,jaumard2016articulatory}). However, these methods are often limited to individual frame analysis of static tongue shapes \cite{xu2016comparative}. In this article, built on previous works on tongue motion prediction, we explore the use of convolutional long short-term memory network (ConvLSTM) in predicting tongue motions with unlabeled ultrasound videos.

\section{Related work}
\label{sec:format}
Predicting tongue motions in ultrasound images remains largely unexplored in speech production research. Previous studies have primarily focused on extracting static tongue shape features from a single ultrasound frame (e.g., \cite{JI201842,hueber2007eigentongue,cai2011recognition}). A recent work utilized 3-dimensional convolutional neural networks (3DCNN) to predict tongue motions in ultrasound videos, showing that future ultrasound frames can be predicted based on past frames \cite{wu2018predicting}. In contrast to traditional convolutional neural networks, 3DCNNs explicitly incorporate the time dimension into the model by convolving the stacked consecutive input frames with 3-dimensional kernels \cite{ji20133d} and therefore are suitbale for processing ultrasound videos. 

Apart from 3DCNN, long short-term memory (LSTM) structure has also been shown to be suitable for modeling long term dynamics of human motions and has been applied to human motion classification and prediction (e.g., \cite{villegas2017decomposing,zhang2017learning}). ConvLSTM has been proposed to specifically solve spatial-temporal sequence prediction problems \cite{xingjian2015convolutional}. In a ConvLSTM, the convolutional modules can capture the vocal tract features of each ultrasound frame at fine granularity, while LSTM cells exploit the auto-correlated properties of tongue movements to infer future frames. The present article describes the first use of ConvLSTMs to predict tongue motions in unlabeled ultrasound data with good performance.

The contributions of the current work can be summarized as follows: 1) We applied ConvLSTM to predict tongue motions in ultrasound image sequences based on past motions, demonstrating that in most cases ConvLSTM network outperforms the 3DCNN adopted in \cite{wu2018predicting}; 2) \cite{wu2018predicting} only investigated predictions of tongue motions in the immediately following frames. However, we also tested the performance of our method in predicting tongue motions in more distant ultrasound frames and ConvLSTM consistently outperforms 3DCNN in multiple quantitative comparisons. 

\section{Convolutional LSTM neural network}
\label{sec:typestyle}
\begin{figure}[ht]
	\centering
	\includegraphics[scale=0.12]{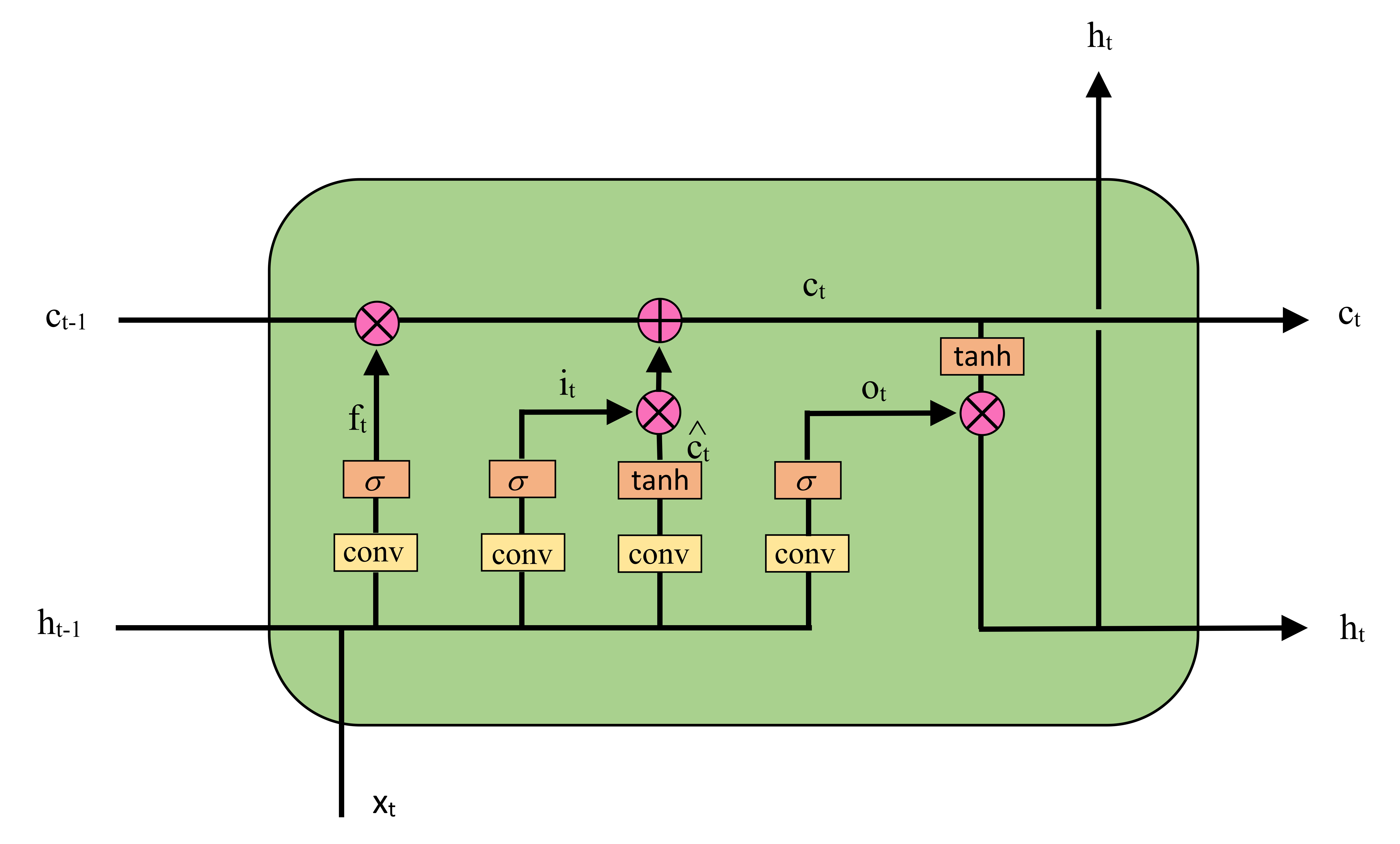}
	\caption{Structure of a convolutional LSTM cell}
	\label{cell}
\end{figure}

\begin{figure}[ht]
	\centering
	\includegraphics[scale=0.13]{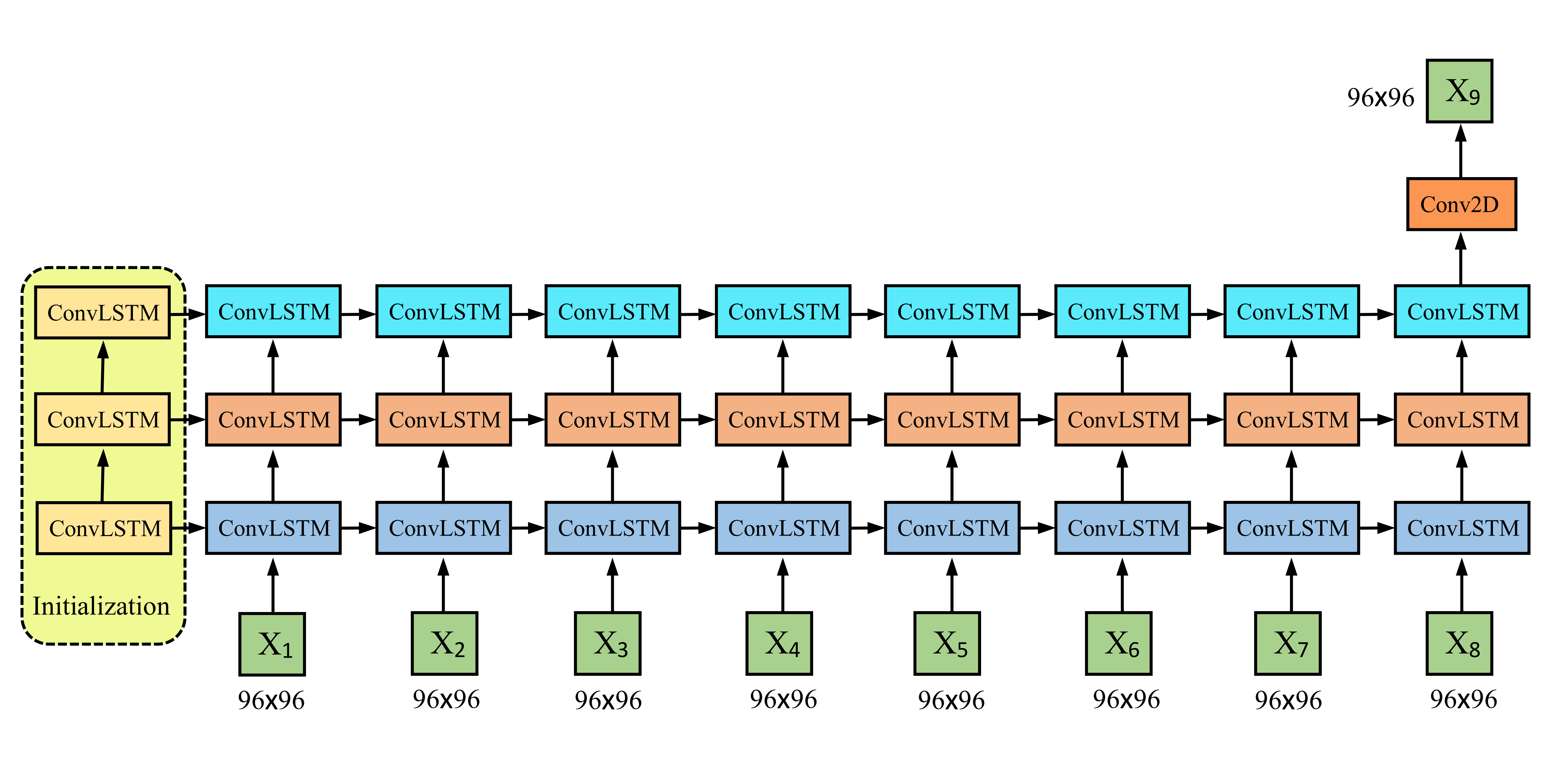}
	\caption{A schematic illustration of the ConvLSTM network}
	\label{convlstm}
\end{figure}

The ConvLSTM is specifically designed for spatial-temporal sequence prediction problems \cite{xingjian2015convolutional}. The convolutional layers can effectively encode the salient visual features of the input images, while LSTM cells can model the temporal dependence across continuous tongue movements with better optimization stability. The structure of a convolutional LSTM cell is illustrated in Fig. \ref{cell}. We adopted a slightly revised version of ConvLSTM \cite{sudhakaran2017learning}, which can be formulated as follows:

\begin{equation}
i_t = \sigma(W_{xi}*X_t+W_{hi}*h_{t-1}+b_i) 
\end{equation}
\begin{equation}
f_t = \sigma(W_{xf}*X_t+W_{hf}*h_{t-1}+b_f) 
\end{equation} 
\begin{equation}
	\widetilde{c}_t = tanh(W_{x\widetilde{c}}*X_t+W_{h\widetilde{c}}*h_{t-1}+b_{\widetilde{c}})
\end{equation}
\begin{equation}
c_t = f_t \circ c_{t-1} + i_t \circ \widetilde{c}_t
\end{equation}
\begin{equation}
o_t = \sigma(W_{xo}*X_t+W_{ho}*h_{t-1}+b_o) 
\end{equation}
\begin{equation}
h_t = o_t \circ tanh(c_t)
\end{equation}
where $X_t$ represents the input tensor, $h_t$ the hidden states, and $c_t$ the cell states. And $i_t$, $f_t$ and $o_t$ are the gate activations at current time step. The ``$\circ$" denotes the Hadamard product operation and ``$*$" denotes convolution operation. The weights ($W$) and biases ($b$) are learned through back propagation during training. In this study, the proposed model was built by stacking multiple ConvLSTM layers to model sequences of ultrasound frames, as illustrated in Fig. \ref{convlstm}.

\section{Dataset}
\label{sec:pagestyle}
Our experiments were based on the same two ultrasound corpora as in \cite{wu2018predicting}. These ultrasound data were recorded as 320x240 pixel mid-sagittal ultrasound tongue images using an acquisition helmet and an ultrasound probe beneath the speaker’s chin. Detailed descriptions can be found in \cite{wu2018predicting}. In all experiments, ultrasound images were resized to 96x96 pixels before further processing.  

The “WSJ0” data were extracted from the Silent Speech Challenge corpus \cite{cai2011recognition}, which contains ultrasound video data at 60 Hz from a male native speaker producing non-vocalized sentences. The total number of training images were more than 700,000 frames of TIMIT sentences, while the number of test images were derived from 35,000 WSJ0 sentences.

“TJU” ultrasound data were collected at Tianjin University by recording a non-native feamle speaker reading a simple training passage (9900 images) and test passage (4800 images) in English. Ultrasound data were recorded at a frame rate of 30 Hz. A region of interest (ROI) containing the tongue was selected before resizing.

Tongue surface contours were extracted automatically with the classic snake algorithm \cite{kass1988snakes} from the TJU dataset. Tongue extraction was only done on the TJU data because the tongue visibility in this data is better than that in the WSJ0 data. These contours represent the position of actual tongue surface in these ultrasound images. A sample can be found in the 3\textsuperscript{rd} image of Fig. \ref{9-cross}.

\section{Experiments}
\begin{figure}[ht]
	\centering
	\includegraphics[scale=0.11]{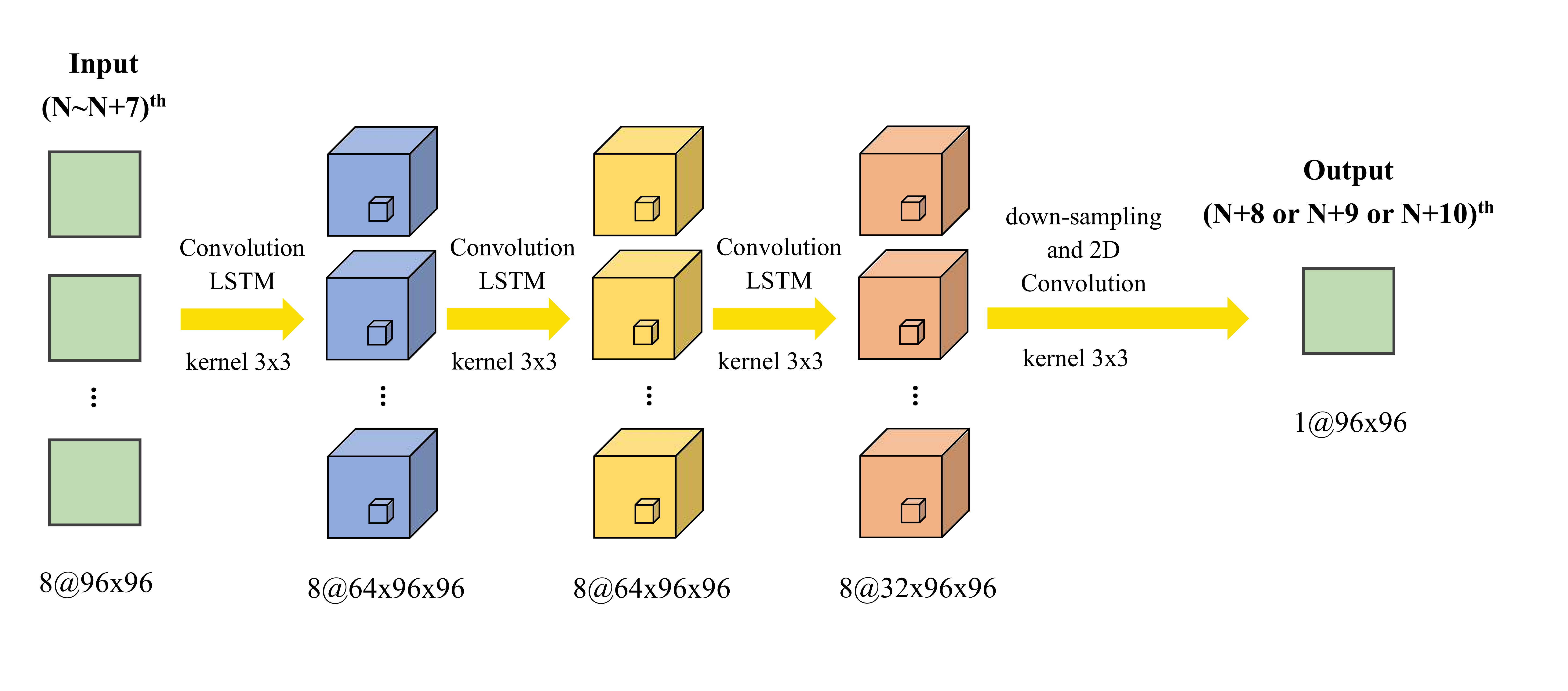}
	\caption{Structure of the ConvLSTM adopted in this study}
	\label{structure}
\end{figure}

A series of experiments were carried out to test the performance of the ConvLSTM architecture in the tongue motion prediction task. The architecture developed is illustrated in Fig.\ref{structure}. It consists of three ConvLSTM layers, and a 2-dimensional convolution layer as the output layer. For all experiments, the inputs were of size 8$\times$96$\times$96, and the outputs of size 1$\times$96$\times$96. We adopted the Adam optimizer with a learning rate of 0.001 and a batch size of 16.

\begin{itemize}
	\item We first compared the performance of ConvLSTM with results in \cite{wu2018predicting} using the same setting. For WSJ0 and TJU data respectively, the input data were generated by stacking 8 temporally consecutive frames together, and the outputs were the immediately following 9\textsuperscript{th} frames. All images from TJU dataset were used in training but only a subset of images from WSJ0 were randomly selected for training due to limited computational resources.
	
	\item For the TJU data, as the snake-based contours were readily available, we also used the 8 consecutive frames to predict the snake-extracted contours in the 9\textsuperscript{th} frame, which was the same as the ``Cross" condition in \cite{wu2018predicting}. In other words, the model is trained to predict the extracted tongue surface contours in the 9\textsuperscript{th} frame. 
	
	\item We also tested the model performance on predicting more distant tongue movements. In these experiments, 8 stacked consecutive ultrasound frames were used to predict the actual tongue image in the following 10\textsuperscript{th} or the 11\textsuperscript{th} frame, that is, future 2 or 3 frames immediately after the input frames. The prediction was evaluated within each ultrasound corpus.
\end{itemize}

\section{Results}
\subsection{Predicting the 9\textsuperscript{th} frame}
In Table \ref{mean-mse}, the performance of the ConvLSTM is compared with other predictiors. As we used the same setting for the TJU data, some of the results obtained in \cite{wu2018predicting} are also listed for comparison. The previous results include the MSE between the actual 9\textsuperscript{th} frames and the output frame obtained from following predictors: the average of the preceding 8 video frames; the 8\textsuperscript{th} frame alone; and the predicted 9\textsuperscript{th} frame given by 3DCNN.  The prediction of ConvLSTM neural network yields a lower MSE value than predictions given by all other methods.

As for the ``Cross" condition, the task is essentially predicting only the tongue surface contour in the 9\textsuperscript{th} frame. The current results may suggest that 3DCNN could be more suitable for this specific subtask of predicting the snake-extracted contours of the 9\textsuperscript{th} frame, and the less optimal performance of ConvLSTM might be due to the influence of other vocal tract features present in the ultrasound frames. However, though the ConvLSTM network does not perform as well as the 3DCNN , its predicted frames still yield far lower MSE than the baseline, which was obtained by computing the MSE between the 8\textsuperscript{th} contour and the 9\textsuperscript{th} contour. These results show that ConvLSTM also has good capacity to isolate the salient tongue contours in ultrasound frames.  

\begin{table}[h]
	\centering 
	\caption{Mean MSE betwen the original frames and the predicted 9\textsuperscript{th} frames by different predictors.}
	\begin{tabular}{|c|c|c|c|}
		\hline
		& WSJ0 & TJU & Cross\\\hline
		Average  & 26.4 & 73.6 \cite{wu2018predicting} & - \\\hline
		8\textsuperscript{th} image  & 22.6 & 40.0 \cite{wu2018predicting} & 279.5 \cite{wu2018predicting} \\\hline
		3DCNN  & 21.3 & 32.6 \cite{wu2018predicting} & \textbf{154.9} \cite{wu2018predicting} \\\hline
		ConvLSTM & \textbf{13.2}  & \textbf{28.1} & 169.2 \\\hline
	\end{tabular}
	\label{mean-mse}
\end{table}

\begin{figure}[ht]
	\centering
	\includegraphics[scale=0.8]{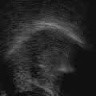}
	\includegraphics[scale=0.8]{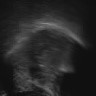}
	\includegraphics[scale=0.8]{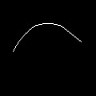}
	\includegraphics[scale=0.8]{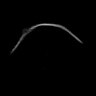}
	\caption{Ultrasound frames and predicted frames in TJU corpus. The left two figures are the 9\textsuperscript{th} frame and the predicted 9\textsuperscript{th} frame respectively. The right two figures are the actual tongue contour and the predicted tongue contour for the same frame.}
	\label{9-cross}
\end{figure}

The quality of the predicted frames is also assessed by the complex wavelet structural similarity index (CW-SSIM) \cite{sampat2009complex}, which has been shown to be particularly robust in measuring similarity between two ultrasound tongue images \cite{xu2016comparative}. In Table \ref{mean-ssim}, the CW-SSIM performance of the convolutional LSTM neural network 9th image predictor is compared to the 3DCNN predictors. The CW-SSIM results again confirm that the predicted frames given by ConvLSTM are more similar to the orginal frames than those given by 3DCNN. Fig. \ref{9-cross} displays sample frames and the corresponding predicted frames from ConvLSTM outputs for visual comparison. Though predicted frames tend to be slightly more blurry than the original frames, these predicted images still contain highly similar tongue shapes and other vocal tract details to those original ultrasound images. 

\begin{table}[h]
	\centering 
	\caption{Mean CW-SSIM between the original frames and the predicted 9\textsuperscript{th} frames by different predictors.}
	\begin{tabular}{|c|c|c|}
		\hline
		& WSJ0 & TJU\\\hline
		3DCNN & 0.907 & 0.912\\\hline
		ConvLSTM & \textbf{0.943} & \textbf{0.952} \\\hline
	\end{tabular}
	\label{mean-ssim}
\end{table}

\begin{figure}[ht]
	\centering
	\includegraphics[scale=0.24]{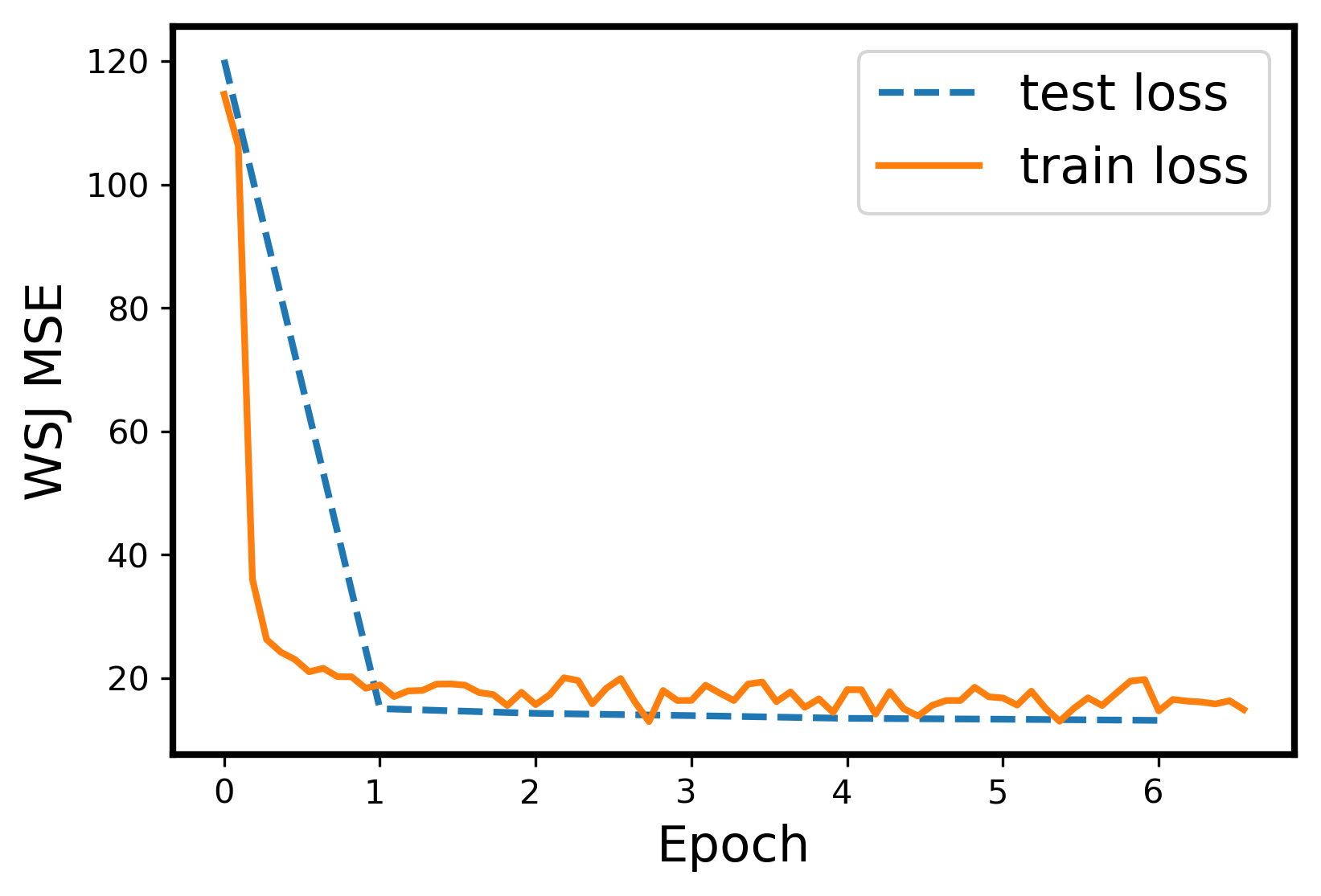}
	\includegraphics[scale=0.24]{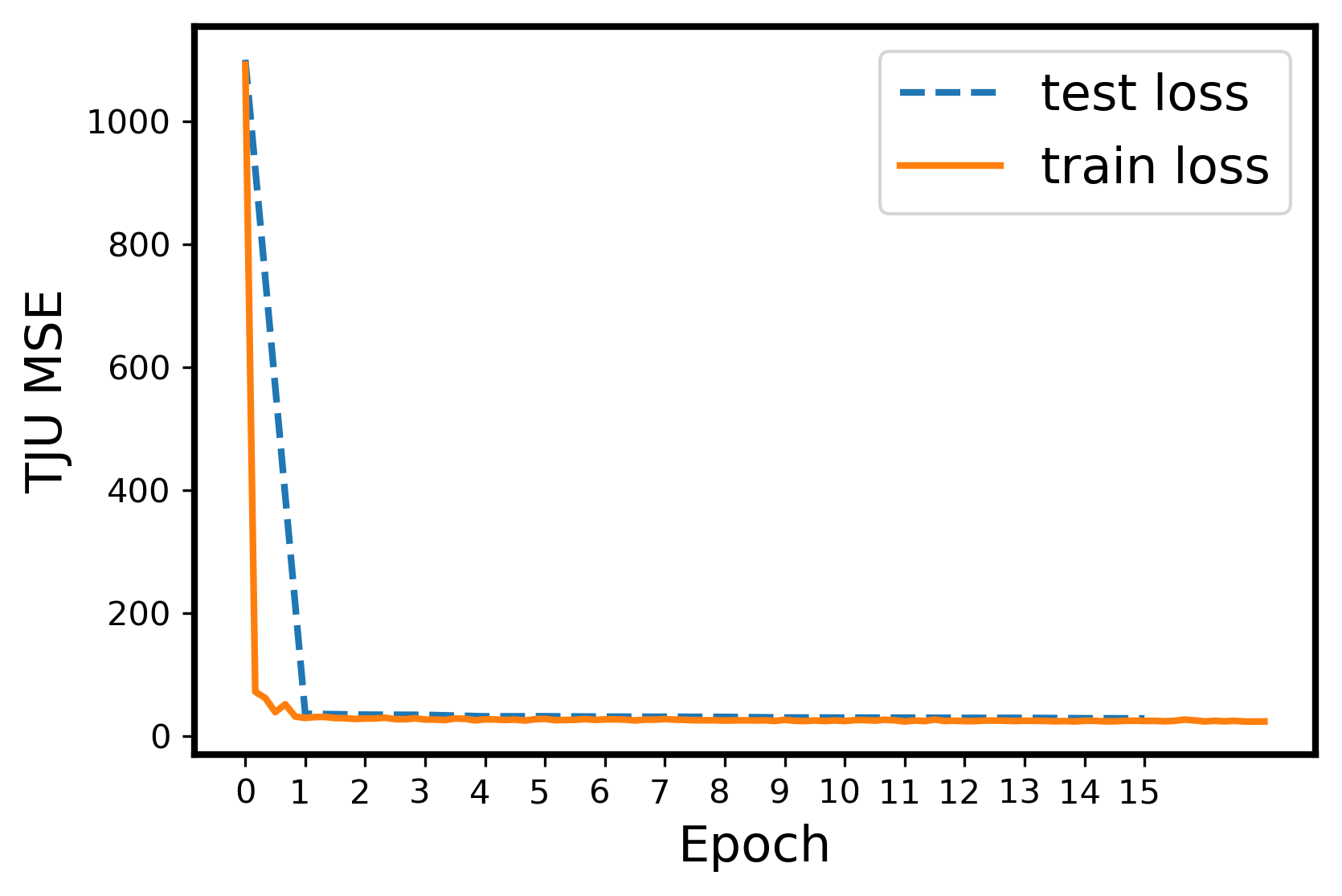}
	\caption{ConvLSTM training and test losses over time for the WSJ0 (left) and TJU Data (right).}
	\label{loss}
\end{figure}

The training and test errors over time for ConvLSTM are displayed in Fig. \ref{loss}. As shown in Fig. \ref{loss}, the ConvLSTM model has high converging speed, high stability and good performance. 

\subsection{Predicting distant tongue motions}
\begin{figure}[ht]
	\centering
	\includegraphics[scale=0.6]{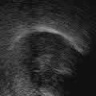}
	\includegraphics[scale=0.6]{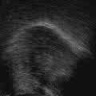}
	\includegraphics[scale=0.6]{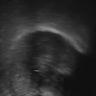}
	\includegraphics[scale=0.6]{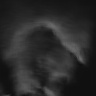}
	\caption{Ultrasound frames and predicted frames in TJU corpus. The left two figures are the 10\textsuperscript{th} and 11\textsuperscript{th} frames respectively. The right two figures are the corresponding predictions by ConvLSTM networks.}
	\label{10-11}
\end{figure}

Table \ref{TJU-lt} and \ref{WSJ0-lt} display the outcomes of distant tongue motion prediction in consecutive ultrasound frames. The results show that both 3DCNN and ConvLSTM performed significantly better than the baseline prediction given by simple averaging in either WSJ0 or TJU corpus, as measured by MSE and CW-SSIM. ConvLSTM consistently yields superior predictions in comparison with 3DCNN under all experimental conditions, further demonstrating its advantage in this particular task. As shown in Fig. \ref{10-11}, the predicted frames closely match the actual frames not only in tongue shapes but also in fine details.

It is also noted that predicting future tongue movements becomes increasingly challenging as the future tongue motion gradually diverges from the current motion. Consequently the prediction quality slightly decreases as the predicted frame gradually moves away from the input frames. The predictive performance on WSJ0 corpus is consistently better than that on TJU corpus, because the WSJ0 corpus was sampled at higher frame rates and tongue motion in a single frame tends to be more autocorrelated with its neighboring frames. 
\begin{table}[h]
	\centering 
	\caption{MSE and CW-SSIM on TJU datasets for different predictors. The indices 10\textsuperscript{th} or 11\textsuperscript{th} indicate which frame is being predicted.}
	\begin{tabular}{|c|c|c|}
		\hline
		Predictor & MSE & CW-SSIM\\\hline
		Average-10\textsuperscript{th} & 100.77 & 0.854 \\\hline
		3DCNN-10\textsuperscript{th} & 48.67 & 0.900\\\hline
		ConvLSTM-10\textsuperscript{th} & \textbf{44.35} & \textbf{0.928} \\\hline
		Average-11\textsuperscript{th} & 116.09 & 0.841 \\\hline
		3DCNN-11\textsuperscript{th} & 66.35 & 0.881\\\hline
		ConvLSTM-11\textsuperscript{th} & \textbf{61.38} & \textbf{0.904} \\\hline
	\end{tabular}
	\label{TJU-lt}
\end{table}

\begin{table}[h]
	\centering 
	\caption{MSE and CW-SSIM on WSJ0 datasets for different predictors. The indices 10\textsuperscript{th} or 11\textsuperscript{th} indicate which frame is being predicted. }
	\begin{tabular}{|c|c|c|}
		\hline
		Predictor & MSE & CW-SSIM\\\hline
		Average-10\textsuperscript{th} & 32.14 & 0.874 \\\hline
		3DCNN-10\textsuperscript{th} & 25.09 & 0.897 \\\hline
		ConvLSTM-10\textsuperscript{th} & \textbf{17.13} & \textbf{0.932} \\\hline
		Average-11\textsuperscript{th} & 35.28 & 0.869 \\\hline
		3DCNN-11\textsuperscript{th} & 28.83 & 0.873 \\\hline
		ConvLSTM-11\textsuperscript{th} & \textbf{20.87} & \textbf{0.913} \\\hline
	\end{tabular}
	\label{WSJ0-lt}
\end{table}

\section{Conclusions}
\label{sec:copyright}
In this study, we demonstrate that ConvLSTMs are capable of predicting future frames in ultrasound videos at the pixel level with high accuracy. ConvLSTM consistently performs better than the 3DCNN and the baseline in predicting future frames, even when tongue motions are distant from the motions in input frames. Though the ConvLSTM shows slightly higher MSE in terms of predicting snake contours directly in comparison with 3DCNN, our method still demonstrates strong capacity to encode variations of tongue shapes in continuous ultrasound frames. 


It is worth noting that the MSE loss function might be guiding the model to predict global vocal tract features in ultrasound frames rather than only the tongue shape, though neural networks do learn to predict the overall tongue shapes accurately. MSE also tends to bias model towards producing blurry images \cite{wu2018predicting}. However, as the tongue shape constitutes the most critical speech-related feature in ultrasound frames, the next step would be to train models that only focus on the tongue shape features. Searching for alternative objective functions might further boost the model performance. 

In the future, it would also be interesting to explore how prediction results vary across speakers and languages, and to what extent tongue motions are predictable. The current work may also have implications for practical applications, such as silent speech interface and articulatory motion synthesis.

\section{ACKNOWLEDGMENT}
This work was supported by the National Grand R\&D Plan(Grant No. 2016YFB1000101). The authors would like to acknowledge useful discussions and helpful suggestions from Zhifeng Gao (Microsoft Asia).
\bibliographystyle{IEEEbib}
\bibliography{ref}

\end{document}